\documentclass[letterpaper]{article} % DO NOT CHANGE THIS
\usepackage[preprint]{aaai2027}  % DO NOT CHANGE THIS
\usepackage[hyphens]{url}  % DO NOT CHANGE THIS
\usepackage{graphicx} % DO NOT CHANGE THIS
\usepackage{natbib}  % DO NOT CHANGE THIS AND DO NOT ADD ANY OPTIONS TO IT
\usepackage{caption} % DO NOT CHANGE THIS AND DO NOT ADD ANY OPTIONS TO IT
\usepackage{amsmath}
\usepackage{amssymb}
\usepackage{multirow}
\usepackage{amsfonts}
\usepackage{xcolor}
\usepackage{makecell}
\usepackage{array}
\usepackage{tikz}
\newcommand{\std}[1]{{\scriptsize$\pm$#1}}

\usepackage{algorithm}
\usepackage{algorithmic}

\usepackage{newfloat}
\usepackage{listings}
\DeclareCaptionStyle{ruled}{labelfont=normalfont,labelsep=colon,strut=off} % DO NOT CHANGE THIS
\floatstyle{ruled}
\newfloat{listing}{tb}{lst}{}
\floatname{listing}{Listing}

\usepackage{booktabs}

\title{ShapeLex: Decoupling Local Shape Symbolization and Global Scale Modeling for Text-Controlled Time Series Generation}
\author{
Subo Wei\textsuperscript{\rm 1},
Jianqi Gao\textsuperscript{\rm 1},
Mingyan Fan\textsuperscript{\rm 1},
Shaorong Xie\textsuperscript{\rm 1},
Xinzhi Wang\textsuperscript{\rm 1},
Yongpeng Dong\textsuperscript{\rm 2}
}

\affiliations{
\textsuperscript{\rm 1}School of Computer Engineering and Science, Shanghai University, Shanghai, China\\
\textsuperscript{\rm 2}Shanghai Institute of Applied Physics, Chinese Academy of Sciences, Shanghai, China
}

\begin{document}

\maketitle

\begin{abstract}
Text-controlled time series generation aims to synthesize, conditioned on
a natural-language description, time series that both follow the semantics
of the description and remain consistent with the distribution of real
data. However, existing paradigms couple semantic understanding and
sequence modeling within a single continuous latent space, lacking both
explicit local semantic anchors and the ability to disentangle global
continuous attributes from local discrete shapes; as a result, key local
structures in the generated series are smoothed, missed, or misplaced.
To address this, we propose Shape Lexicon (ShapeLex), which decouples the
text-to-sequence generation process into two levels: discrete
symbolization of local shapes and continuous modeling of global
attributes. Specifically, ShapeLex generates a series in three steps.
First, it automatically induces a reusable shape vocabulary from the
training data (discrete shape units, e.g., rises, spikes, and sharp
drops), forming a semantically interpretable discrete symbolic space.
Second, an autoregressive generator selects the corresponding shapes from
the vocabulary according to the textual description, adjusts their
attributes such as position and duration to match the local semantics, and
composes them in temporal order into a shape skeleton, thereby achieving
precise alignment from textual semantics to local shapes within the
discrete symbolic space. Finally, a scale head, implemented as a mixture
density network, models and samples the overall level and volatility,
endowing the generated series with a realistic global scale. Experiments
on twelve public datasets, real user-written text, and downstream
forecasting tasks show that the series generated by ShapeLex are more
consistent with the real data distribution than existing methods.
Moreover, the paired supervision data are automatically synthesized from
the vocabulary, so the annotation cost does not grow with dataset size, yielding strong scalability.
\end{abstract}


\section{Introduction}
%------图1 start
\begin{figure}[t]
\centering
\includegraphics[width=\columnwidth]{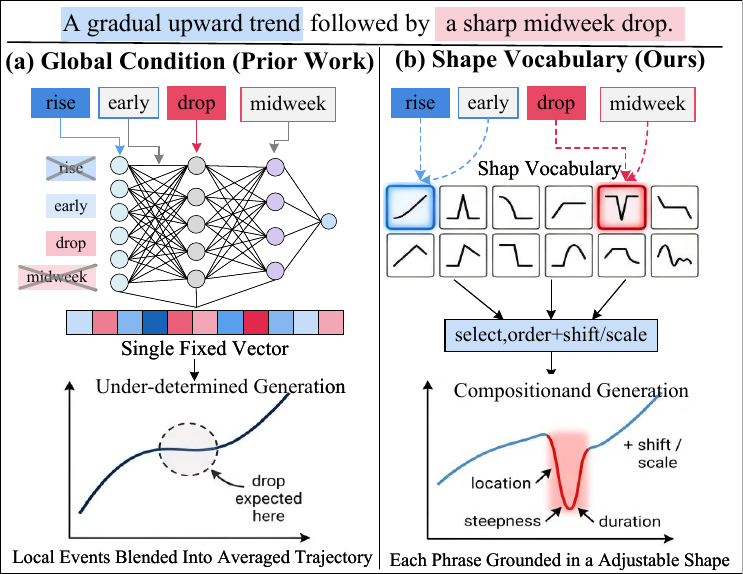}
\caption{Given the same description, (a) a single global condition blends
the specified local events into an averaged trajectory, whereas (b)
ShapeLex grounds each phrase in a shape drawn from a language-addressable
vocabulary and composes them, recovering the specified local structure.}
\label{fig:motivation}
\end{figure}
%------图1 end
Time series generation (TSG) underpins critical decisions in domains such
as energy, transportation, healthcare, and financial risk management~\citep{xu2026timemar,
rousseau2025forging}. In practice, different scenarios impose their own shape requirements on the generated series, and natural language is a direct and intuitive way to express such needs. This gives rise to text-controlled time series generation: synthesizing, conditioned on a natural-language description, series that follow the description while remaining consistent with the distribution of real data.

To turn text into a usable control signal, most existing methods build a
paired text--series representation that aligns linguistic semantics with
the generation process. Yet they largely share one paradigm that couples
semantic understanding and sequence modeling within a single continuous
latent space, compressing the whole description into a global
representation that drives the entire series~\citep{li2025bridge,
huang2025timedp, ge2025t2s}. This coupling has two drawbacks. First, it
offers no explicit local anchor, so a local instruction such as
\textit{``a gradual rise followed by a sharp midweek drop''}
(Figure~\ref{fig:motivation}(a)) can no longer be located or controlled
after compression, and is smoothed, missed, or misplaced in the output.
Second, the global continuous attributes of a series (its overall level
and volatility) are entangled with its local discrete shapes in the same
representation and cannot be separated. As a result, existing methods
capture only the macroscopic trend, while key local structures collapse
into an averaged trajectory.

To address these problems, we propose Shape Lexicon (ShapeLex),
whose core idea is to decouple text-to-sequence generation into two
levels: discrete symbolization of local shapes and continuous modeling of
global attributes (Figure~\ref{fig:motivation}(b)). At the discrete level,
ShapeLex first induces a reusable shape vocabulary from the training data,
abstracting local units such as rises, spikes, and sharp drops into
discrete symbols that form a semantically interpretable space. An
autoregressive generator then selects the corresponding shapes from the
vocabulary according to the description, adjusts attributes such as their
position and duration, and composes them in temporal order into a shape
skeleton, thereby aligning each phrase to a specific local shape rather
than diffusing it over the whole series. At the continuous level, since
the skeleton captures only relative shapes without an absolute scale,
ShapeLex uses a scale head, implemented as a mixture density network
(MDN), to model and sample the overall level and volatility, giving the
series a realistic global scale. Through this division of labor, the
discrete level fixes the local structure while the continuous level
supplies its scale, so ShapeLex renders the local events specified by the
text while keeping the series consistent with real data.

Our main contributions are summarized as follows:
\begin{itemize}
    \item We propose ShapeLex, which decouples text-controlled time series
    generation into discrete symbolization of local shapes and continuous
    modeling of global scale, and introduces a language-addressable symbolic
    level on which text is grounded, alleviating the entanglement of semantics
    and shape in a single continuous latent space.
    
    \item We build a language-addressable shape vocabulary that aligns
    phrases to local shapes, and a scale head that models the global scale, 
    so local structure and scale are generated in a disentangled way.

    \item Extensive experiments on twelve public datasets, real text, and
    downstream forecasting verify the effectiveness of each component and
    the advantage of ShapeLex over strong baselines, with annotation cost
    that does not grow with dataset size.
\end{itemize}

\section{Related Work}
\paragraph{Time series generation.}
Early work synthesizes time series with adversarial and variational
models \citep{yoon2019timegan,jeon2022gtgan,desai2021timevae}, later
joined by vector-quantized priors \citep{lee2023timevqvae} and diffusion
\citep{yuan2024diffusionts}. Recent advances refine the backbone with
selective state spaces \citep{yao2026dimts}, graph-structured latents
\citep{shen2026tsgdiff}, cascaded latent diffusion \citep{shen2026latent},
and prototype guidance \citep{duan2026kprotodiff}, while multi-scale
autoregression is a strong alternative to diffusion \citep{xu2026timemar}.
These methods advance distribution matching, but are unconditional or
driven by low-dimensional signals such as class labels or summary
statistics, and thus cannot follow an open-ended natural-language
description.

\paragraph{Text-conditioned generation.}
Recent methods condition generation on free-form text: domain prompts
from continuous prototypes \citep{huang2025timedp}, multi-agent
bootstrapping of text--series pairs \citep{li2025bridge}, multi-view text
conditioning \citep{gu2025verbalts}, high-resolution text-to-series
diffusion \citep{ge2025t2s}, direct synthesis with large language models
\citep{rousseau2025forging}, and text-enhanced imputation
\citep{xu2026catdiff}. Despite their different routes, all couple text to
the series through a single continuous latent that steers the whole
series, offering no discrete unit that an individual phrase can address;
as a result, local events named in the text are easily smoothed or
misplaced. ShapeLex instead aligns the two modalities at the phrase level
through a shape vocabulary.

\paragraph{Discrete representations of time series.}
Vector quantization yields reusable discrete tokens for generation
\citep{lee2023timevqvae}, abstracted shapes for classification
\citep{wen2024vqshape}, and tokenized series for language-model
forecasting \citep{ansari2024chronos}, alongside codebooks for forecasting
\citep{ma2026recast}, ECG tokens for clinical reasoning
\citep{yang2026heartllm}, and adaptive tokenization \citep{liao2026patk};
in all of these the tokens serve recognition, compression, or prediction
within one modality. ShapeLex instead turns such a vocabulary into a language-addressable interface, through which text composes the generated series.

\section{Proposed Method}
\label{sec:method}

\begin{figure*}[t]
\centering
\includegraphics[width=\textwidth]{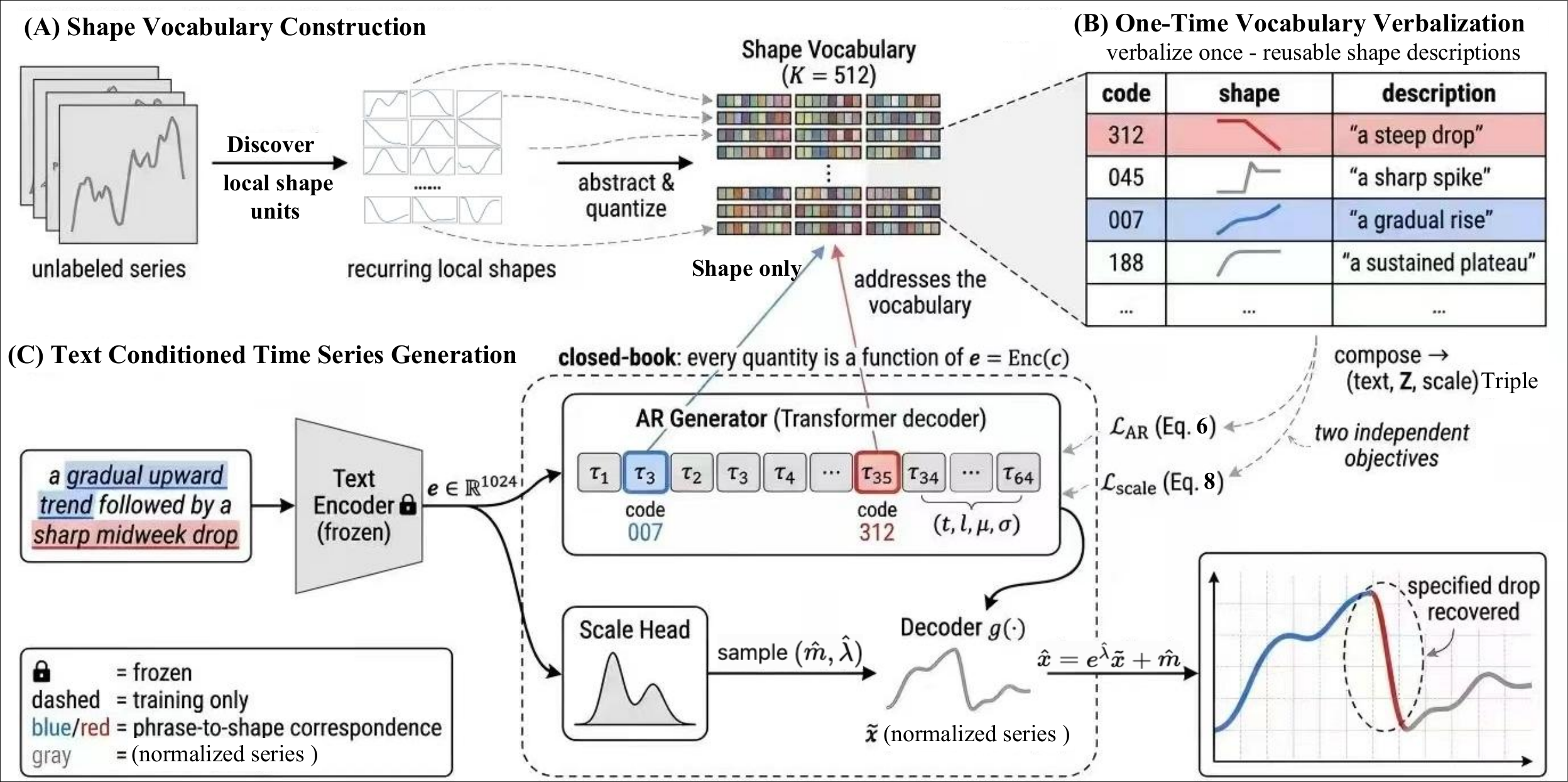}
\caption{Overview of ShapeLex. \textbf{(A) Shape Vocabulary
Construction:} a tokenizer learns discrete shape codes and continuous
local attributes from normalized series. \textbf{(B) One-Time Vocabulary
Verbalization:} shape codes are verbalized once and combined with
instance-specific attribute phrases to synthesize paired supervision.
\textbf{(C) Text-Conditioned Time Series Generation:} text predicts
structured tokens and global scale through separate branches, and the
decoded normalized trajectory is restored to the original scale at the
final step.}
\label{fig:framework}
\end{figure*}

Figure~\ref{fig:framework} presents the three steps of ShapeLex.
First, Shape Vocabulary Construction learns reusable discrete shape codes
and continuous local attributes from normalized time series. Second,
One-Time Vocabulary Verbalization maps each shape code to language and
constructs paired text--token supervision. Finally, conditioned on text,
an autoregressive generator predicts the structured token sequence, while
a scale head models the global level and volatility. The frozen decoder
reconstructs the normalized trajectory from predicted tokens, and the
sampled scale restores the original value range at the final step. The
vocabulary modules are trained and frozen before conditional generation.

Given a description $c$, a frozen text encoder produces
$\mathbf{e}=\operatorname{Enc}(c)\in\mathbb{R}^{1024}$ as the conditioning
signal. It is not decoded into samples or treated as a continuous latent
of the target series; instead, the conditional generator predicts the
structured variables below. For a training window
$\mathbf{x}\in\mathbb{R}^{T}$, we normalize it using its mean $m$ and
standard deviation $v$ as
$\bar{\mathbf{x}}=(\mathbf{x}-m)/(v+\epsilon)$, where $\epsilon$ is a
small stability constant, and resample it from $T{=}96$ to $L{=}512$ by
$\mathbf{u}=\operatorname{Resample}_{T\rightarrow L}(\bar{\mathbf{x}})$.

The normalized series is represented by $N{=}64$ ordered tokens:
\begin{equation}
Z=(\tau_1,\ldots,\tau_N), \qquad
\tau_i=(s_i,t_i,l_i,\mu_i,\sigma_i),
\label{eq:token_definition}
\end{equation}
where $s_i$ denotes the discrete shape identity, $t_i$ and $l_i$ denote
the start and duration, and $\mu_i$ and $\sigma_i{>}0$ represent local
offset and relative amplitude in the normalized domain. We write
$\boldsymbol{\kappa}_i=(t_i,l_i,\mu_i,\sigma_i)$ for the continuous
attributes. Thus, $\mu_i$ and $\sigma_i$ describe local variation without
encoding the absolute scale of the original window. Each token corresponds
to a dynamic interval $\mathcal{I}_i=[t_i,t_i+l_i]$; the token budget is
fixed, whereas locations and durations are predicted and may overlap. We
preserve a deterministic serialization order across tokenization,
verbalization, teacher forcing, and autoregressive inference, so each text
fragment and prediction step refers to the same token position.

\subsection{Shape Vocabulary Construction}
\label{sec:vocabulary}

To obtain reusable local morphology, an analysis tokenizer $A_{\phi}$
extracts
$\{\mathbf{z}_i,t_i,l_i,\mu_i,\sigma_i\}_{i=1}^{N}$ from $\mathbf{u}$,
where $\mathbf{z}_i\in\mathbb{R}^{d_z}$ is a $d_z$-dimensional shape
latent. Only $\mathbf{z}_i$ is quantized into a codebook
$E=\{\mathbf{e}^{s}_{k}\}_{k=1}^{K}$, where $K$ is the codebook size and
$k$ indexes its entries:
\begin{equation}
s_i=
\arg\min_{k\in\{1,\ldots,K\}}
\left\|
\mathbf{z}_i-\mathbf{e}^{s}_{k}
\right\|_2^2 .
\label{eq:shape_quantization}
\end{equation}
where $s_i$ is the selected code index. We set $K{=}512$ and $d_z{=}8$. During training, the decoder receives
the straight-through value
$\widetilde{\mathbf{z}}_i=\mathbf{z}_i+
\operatorname{sg}[\mathbf{e}^{s}_{s_i}-\mathbf{z}_i]$, where
$\operatorname{sg}$ stops gradients. The four attributes bypass the
codebook and remain continuous and sample-specific, so the shape code and
attributes maintain different roles and later receive categorical and
regression supervision, respectively.

The operator $\operatorname{Emb}$ maps each structured token to the
decoder input, and the normalized trajectory is reconstructed as
$
\widehat{\mathbf{u}}
=D_{\psi}\left(
[\operatorname{Emb}(\widetilde{\mathbf{z}}_i,t_i,l_i,\mu_i,\sigma_i)]_{i=1}^{N}
\right).
$
The decoder predicts the normalized trajectory in one pass,
allowing it to coordinate interactions and overlapping
intervals without explicit placement, averaging, or fusion rules.
Importantly, it receives structured token fields rather than a single
sequence-level latent, keeping the shape code and continuous attributes
available for separate prediction and intervention.

The tokenizer, codebook, and decoder are optimized with reconstruction,
codebook, and commitment losses:
\begin{align}
\mathcal{L}_{\mathrm{vocab}}
={}&
\frac{1}{L}\|\mathbf{u}-\widehat{\mathbf{u}}\|_2^2
+\sum_{i=1}^{N}
\left\|
\operatorname{sg}[\mathbf{z}_i]
-\mathbf{e}^{s}_{s_i}
\right\|_2^2
\nonumber\\
&+
\beta
\sum_{i=1}^{N}
\left\|
\mathbf{z}_i-
\operatorname{sg}[\mathbf{e}^{s}_{s_i}]
\right\|_2^2 .
\label{eq:vocabulary_loss}
\end{align}
where $\beta$ controls the commitment term. Continuous attributes are
learned through whole-series reconstruction. After construction,
$A_{\phi}$, $E$, and $D_{\psi}$ are frozen.

\subsection{One-Time Vocabulary Verbalization}
\label{sec:verbalization}

To make the discrete vocabulary addressable by language, each code is
verbalized once. Under a fixed reference setting, code $s$ produces a
canonical normalized waveform $\mathbf{d}_s$. A rule-based descriptor
$\varphi(\mathbf{d}_s)$ summarizes its direction, slope, extrema,
monotonicity, and oscillatory behavior, and a fixed paraphrase set forms
the reusable description family
$D(s)=\operatorname{Para}(\varphi(\mathbf{d}_s))$.
The resulting code-to-language table remains fixed during generator
training.

For each unlabeled training window, the frozen tokenizer provides the
ordered sequence $Z=(\tau_1,\ldots,\tau_N)$. Training-fitted discretizers
$Q_t,Q_l,Q_{\mu},Q_{\sigma}$ convert the corresponding continuous
attributes into phrases, and the fixed template $\operatorname{Compose}$
arranges them in token order:
$c=\operatorname{Compose}
(\{D(s_i),Q_t(t_i),Q_l(l_i),Q_{\mu}(\mu_i),
Q_{\sigma}(\sigma_i)\}_{i=1}^{N})$.

Exact attribute values, raw samples, and global statistics are excluded
from text. This produces training triples
$(c,Z,\mathbf{y}_g)$ with the global-scale target
$\mathbf{y}_g=(m,\log(v+\epsilon))$. Therefore, manual description cost
depends on vocabulary size rather than the number of training windows.
For datasets with human-written reports, the report is directly used as
$c$, while $Z$ and $\mathbf{y}_g$ are extracted from paired series.

\subsection{Text-Conditioned Time Series Generation}
\label{sec:generation}

ShapeLex models the structured local representation and global scale
through separate conditional distributions:
\begin{equation}
p_{\theta,\omega}(Z,\mathbf{y}_g\mid c)
=
p_{\theta}(Z\mid\mathbf{e})
p_{\omega}(\mathbf{y}_g\mid\mathbf{e}).
\label{eq:factorization}
\end{equation}

This factorization prevents window-level mean and volatility from being
absorbed into the local token sequence. The scale branch neither receives
nor modifies shape codes, and the two outputs are combined only after the
normalized trajectory is decoded.

\paragraph{Autoregressive Token Generation.}

The token generator factorizes as
$
p_{\theta}(Z\mid\mathbf{e})
=
\prod_{i=1}^{N}
p_{\theta}(\tau_i\mid\tau_{<i},\mathbf{e})
$.

Under teacher forcing, a Transformer decoder summarizes preceding tokens
and the text condition into a contextual state $\mathbf{h}_i$. Two heads
predict the discrete shape identity and continuous attribute vector:
\begin{equation}
\boldsymbol{\pi}_i
=
\operatorname{softmax}(W_s\mathbf{h}_i),
\qquad
\widehat{\boldsymbol{\kappa}}_i
=
W_{\kappa}\mathbf{h}_i .
\label{eq:ar_heads}
\end{equation}

Both heads share the same autoregressive context but remain in different
prediction spaces. The shape code uses categorical likelihood, while the
attributes use continuous regression, balanced by
$\lambda_{\mathrm{attr}}$:
\begin{equation}
\mathcal{L}_{\mathrm{AR}}
=
\sum_{i=1}^{N}
\left[
-\log \pi_{i,s_i}
+
\lambda_{\mathrm{attr}}
\left\|
\widehat{\boldsymbol{\kappa}}_i-
\boldsymbol{\kappa}_i
\right\|_2^2
\right].
\label{eq:ar_loss}
\end{equation}

\paragraph{Global Scale Modeling.}

The normalized token sequence does not determine a unique absolute level
or volatility. Therefore, we model
$\mathbf{y}_g=(m,\log(v+\epsilon))$ with a mixture density network:
\begin{equation}
p_{\omega}(\mathbf{y}_g\mid\mathbf{e})
=
\sum_{r=1}^{K_s}
\alpha_r(\mathbf{e})
\mathcal{N}
\left(
\mathbf{y}_g;
\boldsymbol{\mu}_r(\mathbf{e}),
\boldsymbol{\Sigma}_r(\mathbf{e})
\right),
\label{eq:scale_mdn}
\end{equation}
where $K_s$ is the number of mixture components, set to $10$; the weights
satisfy $\alpha_r\geq0$ and $\sum_r\alpha_r=1$, and the covariance
matrices remain positive definite. The scale branch is trained with
$\mathcal{L}_{\mathrm{scale}}
=-\log p_{\omega}(\mathbf{y}_g\mid\mathbf{e})$ and shares only the frozen
text condition with the token generator.

\paragraph{Generation and Training.}

At inference, the autoregressive branch predicts $\widehat{Z}$ and the
frozen decoder reconstructs a normalized trajectory. Meanwhile, the scale
branch samples $(\widehat m,\widehat\lambda)$, where
$\widehat\lambda$ is the predicted log-standard-deviation. The final
series is generated as
\begin{equation}
\widehat{\mathbf{x}}
=
\exp(\widehat{\lambda})
\operatorname{Resample}_{L\rightarrow T}
\left(
D_{\psi}(\widehat Z)
\right)
+
\widehat m .
\label{eq:final_generation}
\end{equation}

Training proceeds in three steps: vocabulary construction with
$\mathcal{L}_{\mathrm{vocab}}$, token generation with
$\mathcal{L}_{\mathrm{AR}}$, and global-scale modeling with
$\mathcal{L}_{\mathrm{scale}}$. The text encoder remains frozen
throughout. At inference, text is the only input; no target value or
statistics are available.

\begin{table*}[t]
\centering
\resizebox{\textwidth}{!}{%
\begin{tabular}{cl ccccccccc}
\toprule
& \textbf{Dataset}
& \makecell{\textbf{ShapeLex}\\\textbf{(Ours)}}
& \makecell{TimeMAR\\(WWW 2026)}
& \makecell{BRIDGE\\(ICML 2025)}
& \makecell{T2S\\(IJCAI 2025)}
& \makecell{TimeVQVAE\\(AISTATS 2023)}
& \makecell{GT-GAN\\(NeurIPS 2022)}
& \makecell{TimeVAE\\(arXiv 2021)}
& \makecell{TimeGAN\\(NeurIPS 2019)} \\
\midrule
\multirow{12}{*}{\rotatebox{90}{Marginal Distribution Distance}}
& Electricity & \textbf{0.0066}\std{.0012} & \underline{0.0109}\std{.0020} & 0.220\std{.070} & 0.0355\std{.0043} & 1.763\std{.088} & 2.026\std{.280} & 3.306\std{.044} & 2.443\std{.765} \\
& Solar       & \underline{52.129}\std{2.037} & 53.080\std{4.21} & 375.531\std{.001} & \textbf{50.194}\std{2.615} & 466.174\std{.145} & 476.196\std{17.041} & 365.906\std{6.365} & 460.810\std{14.078} \\
& Wind        & \textbf{0.0370}\std{.0032} & \underline{0.0578}\std{.0049} & 0.316\std{.031} & 0.2552\std{.0231} & 0.777\std{.028} & 0.706\std{.106} & 0.943\std{.008} & 1.115\std{.159} \\
& Traffic     & \textbf{0.0244}\std{.0054} & \underline{0.0246}\std{.0072} & 0.254\std{.034} & 0.0957\std{.0144} & 1.170\std{.028} & 1.311\std{.032} & 0.984\std{.012} & 1.733\std{.137} \\
& Taxi        & \textbf{0.0630}\std{.0088} & \underline{0.0672}\std{.0116} & 0.386\std{.057} & 0.0713\std{.0093} & 0.534\std{.032} & 1.118\std{.157} & 0.697\std{.007} & 1.278\std{.168} \\
& Pedestrian  & \textbf{0.0442}\std{.0068} & \underline{0.0477}\std{.0138} & 0.621\std{.124} & 0.0875\std{.0096} & 1.225\std{.060} & 1.559\std{.117} & 0.777\std{.012} & 1.574\std{.290} \\
& Air         & \textbf{0.0302}\std{.0069} & \underline{0.0320}\std{.0092} & 0.447\std{.112} & 0.0913\std{.0164} & 0.338\std{.012} & 2.828\std{.172} & 1.369\std{.040} & 2.089\std{.618} \\
& Temperature & \textbf{0.0964}\std{.0075} & \underline{0.0978}\std{.0029} & 0.342\std{.010} & 0.1441\std{.0101} & 0.943\std{.035} & 1.165\std{.072} & 2.044\std{.024} & 1.164\std{.110} \\
& Rain        & \textbf{0.1887}\std{.0091} & 0.7141\std{.0325} & 5.340\std{.421} & \underline{0.4361}\std{.0393} & 9.243\std{.122} & 6.473\std{1.207} & 9.134\std{.477} & 10.937\std{4.039} \\
& NN5         & \textbf{0.1445}\std{.0187} & \underline{0.1711}\std{.0275} & 0.591\std{.029} & 0.2155\std{.0259} & 1.424\std{.043} & 2.121\std{.094} & 2.871\std{.045} & 2.758\std{.142} \\
& Fred-MD     & \textbf{0.0363}\std{.0059} & 0.4068\std{.0101} & \underline{0.258}\std{.045} & 0.2607\std{.0365} & 2.932\std{.133} & 4.026\std{.087} & 2.902\std{.215} & 4.028\std{.130} \\
& Exchange    & \textbf{0.3103}\std{.0206} & \underline{0.3165}\std{.0029} & 0.374\std{.053} & 0.3492\std{.0210} & 0.993\std{.058} & 1.355\std{.072} & 1.331\std{.042} & 1.553\std{.122} \\
\midrule
\multirow{12}{*}{\rotatebox{90}{K-L Divergence}}
& Electricity & \textbf{0.0089}\std{.0014} & \underline{0.0093}\std{.0011} & 0.011\std{.010} & 0.2322\std{.0232} & 0.185\std{.018} & 0.415\std{.040} & 0.580\std{.005} & 0.395\std{.121} \\
& Solar       & 0.2328\std{.0072} & 0.2533\std{.0020} & \textbf{0.007}\std{.002} & 2.5605\std{.2048} & 0.726\std{.043} & \underline{0.102}\std{.045} & 0.201\std{.008} & 0.889\std{.288} \\
& Wind        & \textbf{0.0353}\std{.0088} & 0.0672\std{.0049} & \underline{0.067}\std{.030} & 3.7456\std{.4495} & 0.493\std{.081} & 0.511\std{.129} & 0.553\std{.014} & 4.528\std{1.743} \\
& Traffic     & \textbf{0.0109}\std{.0021} & \underline{0.0116}\std{.0040} & 0.013\std{.004} & 0.2902\std{.0261} & 0.145\std{.015} & 1.108\std{.171} & 0.212\std{.006} & 2.134\std{.952} \\
& Taxi        & \textbf{0.0122}\std{.0039} & 0.0300\std{.0090} & \underline{0.013}\std{.009} & 0.0837\std{.0126} & 0.100\std{.014} & 0.663\std{.127} & 0.120\std{.005} & 1.160\std{.651} \\
& Pedestrian  & \underline{0.0374}\std{.0027} & 1.4432\std{.1011} & \textbf{0.011}\std{.009} & 0.1892\std{.0208} & 0.275\std{.021} & 0.347\std{.085} & 0.052\std{.010} & 0.881\std{.436} \\
& Air         & \textbf{0.0083}\std{.0008} & \underline{0.0086}\std{.0017} & 0.022\std{.017} & 0.1073\std{.0172} & 0.017\std{.004} & 0.506\std{.091} & 0.176\std{.016} & 0.588\std{.369} \\
& Temperature & \textbf{0.0188}\std{.0052} & 0.1159\std{.0230} & \underline{0.023}\std{.013} & 0.5432\std{.0706} & 0.980\std{.190} & 2.177\std{.323} & 1.910\std{.076} & 8.775\std{2.511} \\
& Rain        & \underline{0.0067}\std{.0007} & 0.0638\std{.0106} & \textbf{0.006}\std{.001} & 3.8021\std{.2662} & 0.008\std{.002} & 0.462\std{.056} & 0.175\std{.011} & 0.383\std{.089} \\
& NN5         & \underline{0.0126}\std{.0044} & 0.7866\std{.0052} & \textbf{0.010}\std{.008} & 0.9707\std{.0874} & 0.603\std{.107} & 1.372\std{.180} & 1.284\std{.058} & 4.054\std{1.592} \\
& Fred-MD     & \underline{0.3007}\std{.0087} & 4.6959\std{.2194} & \textbf{0.024}\std{.019} & 4.1329\std{.2480} & 0.712\std{.054} & 3.509\std{.299} & 0.376\std{.025} & 5.371\std{1.455} \\
& Exchange    & \textbf{0.0725}\std{.0063} & 4.1141\std{.2881} & \underline{0.083}\std{.056} & 1.9615\std{.2746} & 1.984\std{.836} & 1.583\std{.932} & 2.011\std{.433} & 4.376\std{.664} \\
\midrule
& \makecell[l]{Count(\textbf{best}/\underline{2nd})}
& \textbf{18}/\underline{5} & 0/\underline{11} & \textbf{5}/\underline{5} & \textbf{1}/\underline{0} & 0/0 & 0/\underline{1} & 0/0 & 0/0 \\
\bottomrule
\end{tabular}%
}
\caption{Generation fidelity on twelve univariate benchmarks. ShapeLex,
TimeMAR, BRIDGE, and T2S use the same text-conditioned protocol; the
unconditional methods follow the BRIDGE benchmark and provide general TSG
references.}
\label{tab:main}
\end{table*}
\section{Experiments}
\subsection{Experimental Setup}
\subsubsection{Datasets.}
We evaluate on two groups of data. The first contains twelve univariate
benchmarks from GluonTS \citep{alexandrov2020gluonts} and the Monash
archive \citep{godahewa2021monash}, following the cross-domain setting
of \citet{li2025bridge}. Because these benchmarks do not provide text,
their conditions are synthesized from the frozen shape vocabulary. The
second group contains eight Time-MMD domains \citep{liu2024timemmd}
paired with independently written reports. All series are divided into
non-overlapping windows of length $T{=}96$.
\paragraph{Baselines.}
We compare against the text-conditioned methods TimeMAR
\citep{xu2026timemar}, BRIDGE \citep{li2025bridge}, and T2S
\citep{ge2025t2s}, together with the unconditional generators
TimeVQVAE \citep{lee2023timevqvae}, GT-GAN
\citep{jeon2022gtgan}, TimeVAE \citep{desai2021timevae}, and TimeGAN
\citep{yoon2019timegan}. Following BRIDGE, the unconditional methods
serve as general generation references. ShapeLex and the three
text-conditioned baselines use identical splits, conditions, generated
sample counts, seeds, and metric code; all pairwise significance tests
are restricted to this group.

\paragraph{Protocol parity and no data leakage.}
The vocabulary, independent-condition thresholds, and all trainable
components are constructed from the training split only. Evaluation uses
held-out real windows never seen during vocabulary construction or model
training. At inference, no target-window value or statistic is provided.
Thus Table~\ref{tab:main} measures generation fidelity, while the
independent-condition, real-report, and intervention experiments
separately test whether the results depend on the condition source and
whether text actually controls the output.

\paragraph{Evaluation metrics.}
Following \citet{li2025bridge}, we report Marginal Distribution Distance
(MDD) and K-L divergence for marginal fidelity. We additionally use ACF
distance over lags $1{:}24$ and PSD distance over normalized periodograms
to evaluate temporal structure. Controllability is measured with global
text--output correlations, compositional intervention success, and change
inside versus outside the edited interval. For downstream utility, we use
train-on-synthetic, test-on-real (TSTR) forecasting.
\subsubsection{Implementation Details.}
The text encoder is Qwen3-Embedding-0.6B and remains frozen. The
autoregressive generator is a six-layer Transformer decoder
($d_{\mathrm{model}}{=}512$, eight heads) over $N{=}64$ dynamic tokens.
The shape vocabulary contains $K{=}512$ codes with $d_z{=}8$, and the
global-scale head is a mixture-density network with $K_s{=}10$
components. Unless stated otherwise, results use five random seeds
$\{0,42,123,2024,3407\}$, and multi-dataset audits report the
macro-average of per-dataset seed means.
To assess ShapeLex systematically, we organize the experiments around
three connected questions. First, does ShapeLex generate realistic series
and retain its advantage under vocabulary-derived, tokenizer-independent,
and human-written conditions (Section~\ref{sec:fidelity}). Second, do text
edits and internal token interventions produce localized, factor-specific
responses (Section~\ref{sec:control}). Third, are the generated samples
useful for training real forecasters, and which design choices explain the
gains (Sections~\ref{sec:downstream} and~\ref{sec:ablation}). We begin
with generation quality.
\subsection{Generation Quality and Generalization}
\label{sec:fidelity}
To evaluate whether ShapeLex produces realistic time series rather than
only plausible local patterns, we first compare the generated and real
distributions on the twelve public benchmarks, then change the source of
the condition text, and finally move to independently written reports, so
the evidence progresses from benchmark fidelity to robustness and
natural-language generalization.

\paragraph{ShapeLex matches the target distribution most tightly.}
ShapeLex attains the lowest MDD on eleven of twelve datasets, reducing MDD
by up to $86\%$ over the strongest competitor on heavy-tailed series such
as Fred-MD and Rain. The only exception is Solar, whose magnitudes follow
the non-comparable normalization of \citet{li2025bridge} and are therefore
not directly comparable. It ranks first on seven and second on four
datasets for K-L, giving eighteen best and five second-best results
overall. Holm-adjusted one-sided Wilcoxon tests for MDD give
$p=7.32\times10^{-4}$ against TimeMAR and BRIDGE and $p=0.017$ against T2S.
Because the unconditional methods serve only as general references, these
tests use the reproduced text-conditioned group.
\paragraph{The advantage persists beyond the native condition source.}
The conditions above are derived from the frozen ShapeLex tokenizer. To
test whether this representation-aligned source determines the ranking, we
map region-wise shape statistics to generic phrases using training-split
thresholds, as detailed in Appendix~D.2. All text-conditioned methods are
retrained on the same descriptions for Electricity, Wind, and Traffic, and
neither the condition generator nor the evaluator reads ShapeLex tokens.
\begin{table}[t]
\centering
\small
\begin{tabular}{lcccc}
\toprule
\textbf{Method}
& \textbf{MDD$\downarrow$}
& \textbf{K-L$\downarrow$}
& \textbf{ACF$\downarrow$}
& \textbf{PSD$\downarrow$} \\
\midrule
ShapeLex & \textbf{0.0341} & \textbf{0.0279} & \textbf{0.091} & \textbf{0.103} \\
TimeMAR  & 0.0418 & 0.0362 & 0.106 & 0.109 \\
BRIDGE   & 0.1872 & 0.0754 & 0.163 & 0.181 \\
T2S      & 0.0907 & 0.2183 & 0.128 & 0.142 \\
\bottomrule
\end{tabular}%
\caption{Condition-source robustness under independently constructed descriptions. Lower is better.}
\label{tab:independent_condition}
\end{table}
ShapeLex remains best on all four metrics, with the largest gains on MDD
and K-L ($18.4\%$ and $22.9\%$ over TimeMAR) and smaller but consistent
gains on ACF and PSD ($14.2\%$ and $5.5\%$). The smaller margin than in
Table~\ref{tab:main} indicates that the original condition source
contributes to the gap, while the remaining advantage persists in a
condition space that no evaluated model owns. On the twelve original
benchmarks, ShapeLex also obtains mean ranks of $1.58$ for ACF and $1.67$
for PSD and ranks first on six datasets for each (Appendix~E.2), showing
that its gains extend to serial dependence and spectral structure.
\paragraph{ShapeLex generalizes to independently written reports.}
To test the complete model under realistic text, we use the eight
Time-MMD domains under a closed-book protocol: at inference ShapeLex
receives only the report, and the paired window and its statistics remain
hidden.
\begin{table}[t]
\centering
\resizebox{\columnwidth}{!}{%
\begin{tabular}{cl cccc}
\toprule
& \textbf{Domain} & \textbf{ShapeLex} & TimeMAR & BRIDGE & T2S \\
\midrule
\multirow{8}{*}{\rotatebox{90}{MDD}}
& Energy-MMD        & \textbf{0.1283} & \underline{0.1318} & 0.1548 & 0.1407 \\
& Climate-MMD       & \textbf{0.0564} & \underline{0.0589} & 0.1988 & 0.1205 \\
& Economy-MMD       & \textbf{0.1075} & 0.3998 & 0.2436 & \underline{0.1713} \\
& Agriculture-MMD   & \textbf{0.3722} & 0.4608 & 0.4746 & \underline{0.4186} \\
& Health-Africa-MMD & \textbf{0.1117} & \underline{0.2017} & 0.2488 & 0.2233 \\
& Health-US-MMD     & \underline{0.0766} & \textbf{0.0448} & 0.1924 & 0.1629 \\
& Social-Good-MMD   & \textbf{0.1049} & \underline{0.1220} & 0.2088 & 0.1621 \\
& Traffic-MMD       & \textbf{0.1322} & 0.3005 & \underline{0.2343} & 0.2612 \\
\midrule
\multirow{8}{*}{\rotatebox{90}{K-L}}
& Energy-MMD        & \textbf{0.1671} & 0.2908 & 0.4275 & \underline{0.2829} \\
& Climate-MMD       & \textbf{0.0582} & \underline{0.0788} & 0.9980 & 0.2677 \\
& Economy-MMD       & \textbf{0.2382} & 1.3057 & 1.8607 & \underline{0.4759} \\
& Agriculture-MMD   & \textbf{0.9085} & \underline{1.4543} & 1.5520 & 1.8734 \\
& Health-Africa-MMD & \textbf{0.2533} & 1.3248 & 1.4601 & \underline{0.5821} \\
& Health-US-MMD     & \textbf{0.1217} & \underline{0.1304} & 1.1310 & 0.2036 \\
& Social-Good-MMD   & \textbf{0.1539} & \underline{0.2192} & 1.2840 & 0.3125 \\
& Traffic-MMD       & \textbf{0.4150} & \underline{0.6271} & 1.9360 & 0.8942 \\
\bottomrule
\end{tabular}%
}
\caption{Closed-book generation on Time-MMD domains.}
\label{tab:closedbook}
\end{table}
ShapeLex achieves the lowest MDD on seven of eight domains and the lowest
K-L on all eight, showing that the learned representation transfers beyond
both vocabulary-derived descriptions and the tokenizer-independent
rule-based source. Representative real-report cases comparing value distributions and local
temporal textures are provided in Appendix~D.4.
\subsection{Interventional Controllability and Case Study}
\label{sec:control}
To determine whether the text condition actually changes the
generated series, we intervene at three levels: we first alter
the pairing and encoder to test global response, then edit
individual phrases to measure compositional control, and
finally intervene on the learned tokens to test whether the
internal representation behaves as a local shape lexicon.
\paragraph{Text content controls the generated level.}
On Energy-MMD, report pairing gives $r_{\mathrm{end}}{=}0.363$ and
$r_{\mathrm{mean}}{=}0.563$, whereas shuffling the reports reduces them to
$0.085$ and $0.020$. Replacing Qwen3-Embedding with MiniLM retains
$r_{\mathrm{end}}{=}0.318$, while a random encoder reduces it to $0.012$.
Under the same protocol, BRIDGE obtains $r_{\mathrm{end}}{=}0.110$ and
$r_{\mathrm{mean}}{=}0.218$. The response therefore depends on semantic
content rather than only on the unconditional distribution, although the
correlations indicate measurable rather than perfect numeric control.
\paragraph{Local text edits produce the requested structural changes.}
For each of Electricity, Wind, and Traffic, we build 200 held-out prompt
pairs that differ in exactly one phrase controlling shape, position,
duration, or the order of two events, keeping the random seed and all
decoding settings fixed. A tokenizer-independent waveform detector, with
thresholds fit only on training data, determines whether the requested
event is realized, and locality compares the mean absolute change inside
and outside the edited interval.
\begin{table}[t]
\centering
\resizebox{\columnwidth}{!}{%
\begin{tabular}{lccccc}
\toprule
\textbf{Method}
& \makecell{\textbf{Shape}\\\textbf{success}$\uparrow$}
& \makecell{\textbf{Position}\\\textbf{success}$\uparrow$}
& \makecell{\textbf{Duration}\\\textbf{success}$\uparrow$}
& \makecell{\textbf{Order}\\\textbf{success}$\uparrow$}
& $R_{\mathrm{loc}}\uparrow$ \\
\midrule
ShapeLex & \textbf{0.831} & \textbf{0.742} & \textbf{0.673} & \textbf{0.715} & \textbf{3.28} \\
TimeMAR  & 0.694 & 0.612 & 0.521 & 0.559 & 2.11 \\
BRIDGE   & 0.577 & 0.503 & 0.446 & 0.428 & 1.62 \\
T2S      & 0.652 & 0.531 & 0.487 & 0.544 & 1.79 \\
\bottomrule
\end{tabular}%
}
\caption{Compositional local-edit controllability. 
$R_{\mathrm{loc}}=\Delta_{\mathrm{in}}/(\Delta_{\mathrm{out}}+\epsilon)$
measures how concentrated the change is in the edited interval.Higher is better.}
\label{tab:compositional_control}
\end{table}
ShapeLex reaches $83.1\%$, $74.2\%$, $67.3\%$, and $71.5\%$ success for
shape, position, duration, and event-order edits, exceeding TimeMAR by
$13$--$16$ percentage points. Its locality ratio is $1.55\times$ higher,
showing that the semantic edit is more concentrated in the intended
interval rather than altering surrounding regions.
\paragraph{The learned tokens exhibit local and factor-specific effects.}
To probe the internal tokens directly, we hold all other tokens fixed and
either replace one shape code or perturb one of
$(t_i,l_i,\mu_i,\sigma_i)$.Code consistency measures whether repeated uses
of a code share a waveform family; code-swap and attribute success measure
the intended response; shape retention tests whether continuous edits
preserve morphology; and
$R_{\mathrm{tok}}=\Delta_{\mathcal I_i}/(\Delta_{\overline{\mathcal I_i}}
+\epsilon)$ measures localization. A continuous-latent representation
clustered into $K{=}512$ groups serves as the interacting-slot control.
\begin{table}[t]
\centering
\resizebox{\columnwidth}{!}{%
\begin{tabular}{lccccc}
\toprule
\textbf{Representation}
& \makecell{\textbf{Code}\\\textbf{consistency}$\uparrow$}
& \makecell{\textbf{Swap}\\\textbf{success}$\uparrow$}
& \makecell{\textbf{Attribute}\\\textbf{success}$\uparrow$}
& \makecell{\textbf{Shape}\\\textbf{retention}$\uparrow$}
& $R_{\mathrm{tok}}\uparrow$ \\
\midrule
ShapeLex
& \textbf{0.803} & \textbf{0.726} & \textbf{0.694} & \textbf{0.847} & \textbf{2.94} \\
Continuous latent + $K$-means
& 0.638 & 0.571 & 0.622 & 0.703 & 1.81 \\
\bottomrule
\end{tabular}%
}
\caption{Direct token-level locality and discrete--continuous disentanglement. Higher is better.}
\label{tab:token_intervention}
\end{table}
ShapeLex improves code consistency from $0.638$ to $0.803$ and code-swap
success from $0.571$ to $0.726$. Attribute success rises from $0.622$ to
$0.694$ and shape retention from $0.703$ to $0.847$, supporting the
intended division between discrete morphology and continuous temporal or
affine attributes; the smaller gap on attribute success is expected, since
both representations retain continuous fields. Its locality ratio is $1.62\times$ that of the continuous baseline
($2.94$ versus $1.81$), supporting the view that ShapeLex operates as a soft, context-dependent local lexicon rather than strict token independence.
\begin{figure}[t]
\centering
\includegraphics[width=\columnwidth]{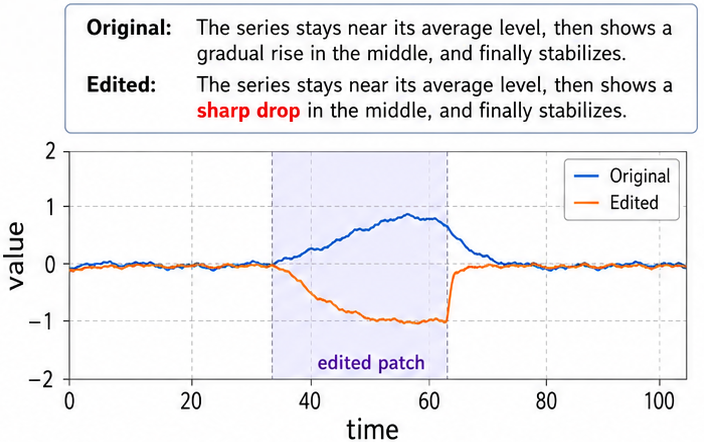}
\caption{Localized response to a shape edit under a fixed seed and
decoding settings.}
\label{fig:local_edit}
\end{figure}

\paragraph{Case study of a localized semantic edit.}
Figure~\ref{fig:local_edit} changes only ``a gradual rise'' to ``a sharp
drop,'' fixing the temporal wording, random seed, and decoding settings.
ShapeLex reverses the slope within the target interval while leaving the
surrounding trajectory largely unchanged, visualizing the concentrated
response quantified above. Additional localized edit cases are provided
in Appendix~F.4.

\subsection{Downstream Utility}
\label{sec:downstream}
To verify that the generated samples are not only statistically faithful
but also useful for learning, we train the same one-step GRU on an equal
number of windows generated by each method and evaluate it on the shared
held-out real test set. All synthetic training sets share the same epochs, hyperparameters, and
model selection; TRTR uses the same architecture trained on the real
split.
\begin{table}[t]
\centering
\small
\begin{tabular}{cl ccc}
\toprule
& \textbf{Dataset} & \textbf{ShapeLex} & BRIDGE & TRTR \\
\midrule
\multirow{3}{*}{\rotatebox{90}{CORR}}
& Electricity & \textbf{0.988} & 0.962 & \textit{0.990} \\
& Wind        & \textbf{1.000} & 0.987 & \textit{1.000} \\
& Traffic     & \textbf{0.922} & 0.851 & \textit{0.928} \\
\midrule
\multirow{2}{*}{\rotatebox{90}{MAE}}
& Energy-MMD  & \textbf{0.067} & 0.073 & \textit{0.055} \\
& Traffic-MMD & \textbf{0.836} & 1.607 & \textit{0.333} \\
\bottomrule
\end{tabular}%

\caption{Downstream utility under train-on-synthetic, test-on-real. Higher is better for CORR, lower for MAE.}
\label{tab:downstream}
\end{table}
\paragraph{Generated samples retain forecast-relevant information.}
ShapeLex nearly matches the TRTR reference on Electricity, Wind, and
Traffic and exceeds BRIDGE on both real-text domains, reducing MAE from
$0.073$ to $0.067$ on Energy-MMD and from $1.607$ to $0.836$ on
Traffic-MMD. Together with the ACF and PSD results, this indicates that
ShapeLex preserves predictive temporal dynamics rather than only matching
marginal value distributions.
\subsection{Ablation and Analysis}
\label{sec:ablation}
To identify which design choices produce the preceding gains, we remove
each component while preserving the data, text conditions, training
budget, seeds, and evaluation code. These ablations test component
necessity, while the direct interventions in
Table~\ref{tab:token_intervention} separately test token semantics.
\begin{table}[t]
\centering
\resizebox{\columnwidth}{!}{%
\begin{tabular}{lccc}
\toprule
\textbf{Variant}
& \makecell{\textbf{Electricity}\\\textbf{MDD$\downarrow$/K-L$\downarrow$}}
& \makecell{\textbf{Wind}\\\textbf{MDD$\downarrow$/K-L$\downarrow$}}
& \makecell{\textbf{Traffic}\\\textbf{MDD$\downarrow$/K-L$\downarrow$}} \\
\midrule
Continuous latent
& 0.2087/0.0124 & 0.1426/0.0815 & 0.1182/0.0437 \\
No local normalization
& 0.0185/0.0157 & 0.0618/0.0724 & 0.0476/0.0289 \\
No local attributes
& 0.0318/0.0276 & 0.0847/0.1042 & 0.1080/0.1250 \\
Point scale regression
& 0.0112/0.0415 & 0.0540/0.2450 & 0.0368/0.0684 \\
No factorization
& 0.0684/0.0931 & 0.1193/0.1837 & 0.0915/0.1428 \\
\midrule
ShapeLex
& \textbf{0.0066/0.0089} & \textbf{0.0370/0.0353} & \textbf{0.0244/0.0109} \\
\bottomrule
\end{tabular}%
}
\caption{Structural ablations on three representative datasets.}
\label{tab:ablation}
\end{table}
\begin{figure}[t]
\centering
\small
\includegraphics[width=\columnwidth]{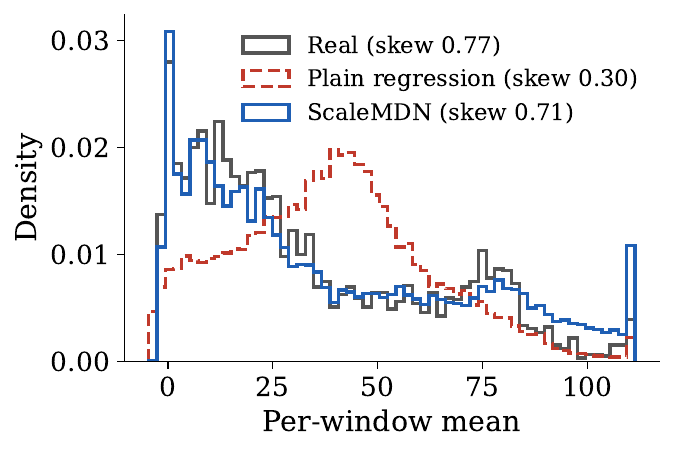}
\caption{Window-level scale distributions of held-out real data and
outputs from point regression and the MDN.}
\label{fig:scale_dist}
\end{figure}
\paragraph{Probabilistic scale modeling preserves cross-window variation.}
The point-regression variant predicts a single scale per condition and
contracts generated windows toward a narrow range. As shown in
Figure~\ref{fig:scale_dist}, this under-represents the spread and tails of
the held-out scale distribution, while the mixture-density head recovers a
broader distribution closer to the real data. This explains why replacing
the MDN with point regression raises Wind K-L from $0.0353$ to $0.2450$
while its effect on MDD is smaller: K-L is more sensitive to lost spread
and tail mass.
\paragraph{The discrete--continuous factorization is necessary.}
Removing vector quantization raises Electricity MDD from $0.0066$ to
$0.2087$. Removing the continuous attributes changes Traffic MDD/K-L from
$0.0244/0.0109$ to $0.1080/0.1250$, showing that a discrete shape identity
alone cannot specify placement and local affine variation. Removing local
normalization or the full local/global factorization also degrades both
metrics on all three datasets. Together with Figure~\ref{fig:scale_dist},
these results show that discrete morphology, continuous token attributes,
and probabilistic global scale make complementary contributions.
\paragraph{A moderate token budget provides the best trade-off.}
Across Electricity, Wind, and Traffic, $N{=}32$ increases mean MDD by
$40.5\%$ relative to $N{=}64$, whereas $N{=}96$ reduces it by only $3.5\%$
at higher computation. We therefore use $N{=}64$. Tokenizer reconstruction
and token-budget sensitivity are detailed in Appendices~C.2 and H.1,
while temporal coverage and overlap are defined in Appendix~H.2.

\section{Conclusion}
We presented ShapeLex, which decouples text-controlled time series
generation into two levels: discrete symbolization of local shapes and
continuous modeling of global scale. A language-addressable shape
vocabulary grounds each phrase on a concrete, nameable shape, while a
mixture density network restores the overall level and volatility that
normalization removes. Because supervision is synthesized from the
vocabulary rather than annotated by hand, its cost stays constant as the
data grows. Across twelve public benchmarks, real human-written reports,
and downstream forecasting, ShapeLex matches the real distribution more
tightly than prior methods and lets the text control local structure
directly, instead of diffusing it over the whole series. Enriching the
vocabulary with rarer shapes and extending it to the multivariate setting
are promising directions for future work.

\bibliography{aaai2027}

\end{document}